\documentclass[10pt,twocolumn,letterpaper]{article}

\usepackage{wacv}              

\usepackage[accsupp]{axessibility}  

\definecolor{wacvblue}{rgb}{0.21,0.49,0.74}
\usepackage[pagebackref,breaklinks,colorlinks,allcolors=wacvblue]{hyperref}

\usepackage{multirow}

\def\wacvPaperID{2532} 
\def\confName{WACV}
\def\confYear{2026}

\title{Multi-Grained Text-Guided Image Fusion\\for Multi-Exposure and Multi-Focus Scenarios}

\author{Mingwei Tang$^1$ \ \ \ \ Jiahao Nie$^{2,\dagger}$ \ \ \ \ Guang Yang$^1$ \ \ \ \ Ziqing Cui$^3$ \ \ \ \ Jie Li$^1$\\
$^1$Xidian University \ \ $^2$Nanyang Technological University \ \ $^3$Xi'an University of Technology\\
{\tt\small mingwei.tang01@outlook.com \ \ \ jiahao007@e.ntu.edu.sg}
}

\begin{document}
\maketitle
\def\thefootnote{$\dagger$}\footnotetext{Corresponding author}\def\thefootnote{\arabic{footnote}}
\begin{abstract}
Image fusion aims to synthesize a single high-quality image from a pair of inputs captured under challenging conditions, such as differing exposure levels or focal depths. A core challenge lies in effectively handling disparities in dynamic range and focus depth between the inputs. With the advent of vision–language models, recent methods incorporate textual descriptions as auxiliary guidance to enhance fusion quality. However, simply incorporating coarse-grained descriptions hampers the understanding of fine-grained details and poses challenges for precise cross-modal alignment. To address these limitations, we propose \textbf{M}ulti-grained \textbf{T}ext-guided \textbf{I}mage \textbf{F}usion (MTIF), a novel fusion paradigm with three key designs. First, it introduces multi-grained textual descriptions that separately capture fine details, structural cues, and semantic content, guiding image fusion through a hierarchical cross-modal modulation module. Second, it involves supervision signals at each granularity to facilitate alignment between visual and textual features and enhance the utility of auxiliary text. Third, it adopts a saliency-driven enrichment module to augment training data with dense semantic content, further strengthening the cross-modal modulation and alignment. Extensive experiments show that MTIF consistently outperforms previous methods on both multi-exposure and multi-focus image fusion tasks.
\end{abstract}
    
\section{Introduction}
Image fusion combines complementary information from multiple inputs to generate an output image with enhanced and more informative details, thereby benefiting both human visual perception and machine-based analysis~\cite{liu2022target,xu2022multi,zhang2021deep}. Multi-exposure image fusion (MEF) and multi-focus image fusion (MFF) are two pivotal tasks in this domain. Specifically, MEF addresses limited dynamic range by integrating images acquired under varying exposure settings~\cite{xu2022multi}, and MFF combines images focused on different depth regions to overcome the limitations of shallow depth of field~\cite{zhang2021deep}. These two tasks are widely recognized as fundamental in image fusion, playing a pivotal role in advancing visual quality~\cite{zhang2021image}. In recent years, an increasing body of research has been devoted to developing image fusion algorithms targeting these challenges~\cite{shen2012qoe,yang2019multiexposure,xu2023unsupervised,liu2023holoco,li2013image_dynamic_scenes,hu2023zmff,zhang2024exploit}.

\begin{figure}[t]
    \centering
    \includegraphics[width=\linewidth]{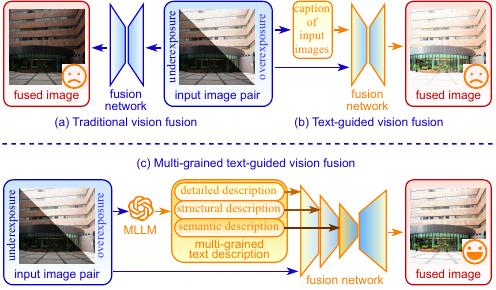}
    \vspace{-6mm}
    \caption{Unlike existing image fusion approaches that either solely rely on visual features as in (a) or simply introduce coarse-grained textual auxiliaries as in (b), the proposed multi-grained textual guidance framework introduces multi-grained textual information that captures fine details, structural cues, and semantic content to facilitate high-quality image fusion as shown in (c).}
    \vspace{-1mm}
    \label{fig:intro}
\end{figure}

As illustrated in Fig.~\ref{fig:intro}(a), initial image fusion methods generally operate by extracting representative features from source images and integrating them to generate a relatively high-quality fusion output~\cite{liu2017multi,ram2017deepfuse,zhang2020ifcnn,xu2020u2fusion,xu2020mef,zhang2021mff,li2024crossfuse}. Despite their success in image fusion, these methods primarily rely on shallow visual cues at the pixel level, while neglecting the deeper semantic information of images, thereby hindering effective modeling of contextual cues~\cite{Zhao_2024_ICML}. As a result, they often lead to the noticeable loss of fine details and structural distortions in semantically critical regions of the fusion images~\cite{zhang2024mrfs}.

With the advent of vision–language models, recent promising methods have introduced vision-language paradigms that leverage textual descriptions to provide high-level semantic guidance for modulating visual feature fusion (refer to Fig.~\ref{fig:intro}(b))~\cite{li2023text,yi2024text,wang2025multi,cheng2025textfusion}. Compared with purely vision-based approaches, these text-guided frameworks can comparatively better preserve fine details, structural integrity, and semantic consistency in fusion images~\cite{zhang2024text}. Despite these advances, existing approaches often rely on coarse-grained textual descriptions, which limit their ability to capture multi-level semantic information, thereby constraining precise modulation of visual features across semantic levels. Consequently, the preservation of fine details and structural consistency in complex, semantically rich regions remains challenging. 

To mitigate these limitations, we propose \textbf{M}ulti-grained \textbf{T}ext-guided \textbf{I}mage \textbf{F}usion (\textbf{MTIF}), a novel framework for MEF and MFF tasks. Specifically, we employ a large language model to generate multi-grained textual descriptions that separately capture fine details, structural cues, and semantic content from input image pairs (refer to Fig.~\ref{fig:intro}(c)). These hierarchical textual features are then fused with visual features through a text-guided visual modulation module, effectively guiding the fusion process and enhancing both detail completeness and semantic consistency in the output images. To bridge the inherent semantic gap between visual and textual modalities and promote effective cross-modal alignment, we further design a multi-grained supervision strategy that not only aligns the fusion output with the input sequence, but also decodes intermediate cross-modal features at multiple semantic levels into the image space and enforces their consistency with the source images. This hierarchical supervision progressively narrows the semantic gap, enabling textual guidance to more effectively modulate visual features across different semantic levels. Moreover, we introduce a semantic-driven enrichment module that segments the input images into multiple salient regions based on object detection, which increases the semantic density and structural diversity of the training data, thereby maximizing the effectiveness of the multi-grained textual guidance mechanism.

\noindent Our contributions can be summarized in three aspects:
\begin{itemize}
\item[$\bullet$] We introduce multi-grained textual descriptions for MEF and MFF tasks, which provide additional semantic information to enhance detail completeness and semantic consistency of the results.
\item[$\bullet$] We introduce a multi-grained cross-modal supervision strategy to better align and integrate information across visual and textual modalities and a semantic-driven enrichment module to increase the semantic density and structural diversity of the training data.
\item[$\bullet$] Extensive experiments on multiple benchmarks demonstrate the effectiveness of our proposed method.
\end{itemize}

\section{Related work}
\subsection{Multi-Exposure Image Fusion}
Multi-exposure image fusion (MEF) aims to synthesize a high-quality image from a sequence of low dynamic range images with varying exposure levels~\cite{xu2022multi}. Traditional MEF approaches can be categorized into spatial-domain and transform-domain methods. Spatial-domain methods~\cite{song2011probabilistic,ma2015multi,ma2017multi,ma2017robust,li2020fast,ulucan2021multi,shen2012qoe} typically rely on hand-crafted weights derived from low-level cues, while transform-domain methods~\cite{li2013image,nejati2017fast,li2017detail,wang2019detail,wang2016multi,yang2019multiexposure} often utilize multi-scale or frequency-based representations to enhance robustness~\cite{shen2014exposure,martorell2019ghosting}. However, both categories rely heavily on manually designed features and fusion rules, which tend to produce artifacts near region boundaries and exhibit limited generalization in complex scenes~\cite{xu2022multi}. Recent advances employ deep learning to adaptively extract and fuse features, and have achieved significant improvements over traditional methods~\cite{ram2017deepfuse,zhang2020ifcnn,xu2020mef,liu2023holoco,xu2023unsupervised}. Nevertheless, most existing models focus primarily on visual information and often overlook deeper text-level semantic information beyond vision. This constrains their ability to capture and align rich semantic cues, and preserve texture details~\cite{Zhao_2024_ICML}.

\subsection{Multi-Focus Image Fusion}
Multi-focus image fusion (MFF) aims to generate an all-in-focus image by integrating source images captured with different focal settings~\cite{zhang2021deep}. Traditional methods are broadly classified into spatial-domain and transform-domain approaches. Spatial-domain methods~\cite{li2001combination,bai2015quadtree,zhang2017boundary,liu2015multi,li2013image_dynamic_scenes} perform pixel-level fusion based on local focus measures, while transform-domain methods~\cite{petrovic2004gradient,phamila2013low,yang2009multifocus} preserve fine details by selecting coefficients. However, these approaches rely on hand-crafted focus metrics and fusion rules, leading to artifacts and poor generalization in complex scenes~\cite{liu2020multi}. With the advancement of deep learning, recent methods achieve remarkable progress by leveraging neural networks~\cite{liu2017multi,zhang2021mff,hu2023zmff}. Nonetheless, most existing methods still fall short in leveraging high-level semantic cues from textual information~\cite{Zhao_2024_ICML}.

\subsection{Vision-Language Models for Image Fusion}
In recent years, vision-language models (VLMs)~\cite{radford2021learning,li2022blip,li2023blip,liu2023visual,wu2024visionllm} have emerged as a prominent research focus, achieving remarkable success across a range of vision tasks~\cite{wu2024general,wang2024llm,agenticir,sun2024coser}. Motivated by their powerful multi-modal representation capabilities, an increasing number of studies have explored their integration into image fusion frameworks~\cite{li2023text,wang2025multi,cheng2025textfusion}. Text-IF~\cite{yi2024text} introduces an interactive fusion framework that leverages semantic text guidance to adaptively modulate the fusion process under different degradation conditions. FILM~\cite{Zhao_2024_ICML} explores the use of pretrained language models to generate global semantic priors, guiding multi-modal fusion. These methods rely on single-level textual features and fail to fully leverage the hierarchical information in both vision and language. This limitation reduces the completeness and consistency of detailed textures and semantic information in the fusion outputs. To address these challenges, we propose a novel multi-grained textual guidance framework for image fusion, which leverages detailed, structural, and semantic-level descriptions to modulate visual features at corresponding levels, thereby enhancing semantic consistency and detail preservation in fusion images.

\begin{figure}[t]
    \centering
    \vspace{-2mm}
    \includegraphics[width=\linewidth]{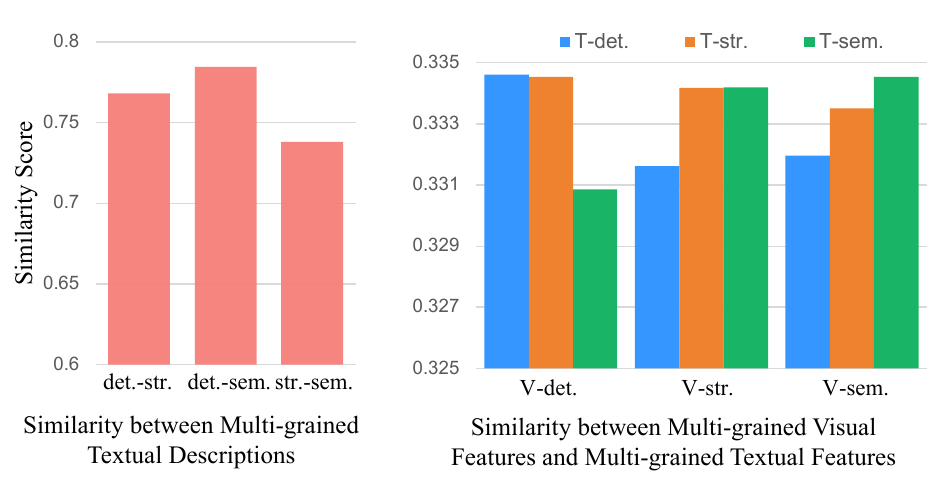}
    \vspace{-2mm}
    \caption{Similarity analysis of multi-grained textual and visual features. \textbf{Left}: Similarity across textual descriptions at the detail, structure, and semantic levels. The relatively low scores imply that these descriptions are semantically complementary, providing richer information than a single-grained textual description. \textbf{Right}: Cross-modal similarity analysis between textual and visual features at multiple levels. The findings indicate that same-level similarities are higher than cross-level similarities, which inspires the multi-grained modulation design of MTIF.}
    \vspace{-2mm}
    \label{fig:sim}
\end{figure}
\begin{figure*}[t]
    \centering
    \vspace{-7mm}
    \includegraphics[width=\linewidth]{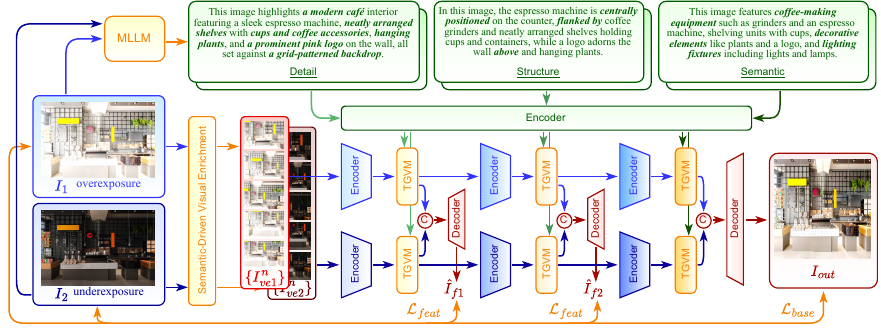}
    \vspace{-8mm}
    \caption{Architecture of the proposed MTIF, which synthesizes a single image from input multi-exposure or multi-focus pairs. First, we utilize a multi-modal large language model to generate multi-grained textual descriptions that capture detail-level, structure-level, and semantic-level information from the input images. Second, we modulate the multi-scale visual features with the corresponding textual features through a text-guided visual modulation module. Moreover, we introduce a multi-grained supervision strategy, which progressively bridges the cross-modal semantic gaps and enables effective feature modulation across all semantic levels. Additionally, we involve semantic-driven visual enrichment to enhance both the diversity and density of the training data.}
    \vspace{-1mm}
    \label{fig:model}
\end{figure*}

\section{Preliminary and Motivation}
Vision-language models (VLMs) have demonstrated impressive capabilities in understanding visual and textual content. Leveraging the rich semantic priors embedded in textual descriptions has emerged as a promising research direction for image fusion~\cite{cheng2025textfusion}. Nevertheless, the prior method simply involves a single-grained textual description as auxiliary, which overlooks the dense fine-grained details~\cite{Zhao_2024_ICML}. To mitigate this limitation, our method employs GPT-4o~\cite{blog2024hello}, a state-of-the-art multi-modal large language model (MLLM), to generate textual descriptions of the input image. Specifically, we prompt GPT-4o~\cite{blog2024hello} to generate multi-grained textual descriptions covering fine details, structural cues, and high-level semantics. Subsequently, we utilize BLIP-2~\cite{li2023blip}, a lightweight and effective vision-language model, to extract textual features aligned with different levels of visual features. These textual features are used to modulate visual features at multiple semantic levels, enhancing both semantic consistency and detail preservation in the fusion outputs.

To validate the motivation for multi-grained textual guidance in image fusion, we first compute the pairwise similarity between the detail-level, structure-level, and semantic-level textual descriptions. As shown in Fig.~\ref{fig:sim} (Left), the similarity between texts of different granularity is moderate, clearly indicating that these descriptions are complementary in semantic content rather than entirely redundant or identical. This suggests that multi-grained textual descriptions introduce diverse semantic cues and provide comprehensive guidance for image fusion. Furthermore, we analyze the cross-modal similarity between textual features and visual features at different levels. As shown in Fig.~\ref{fig:sim} (Right), text features describing fine details exhibit the highest alignment with low-level visual features. Similarly, text features focused on structural cues align well with mid-level features, and those capturing semantic content show the strongest alignment with high-level features. These observations strongly validate the high correspondence between language and vision at corresponding semantic levels, motivating us to design a multi-grained textual guidance mechanism that injects level-specific semantic priors into corresponding visual features.  
\section{Method}
We describe the proposed \textbf{M}ulti-grained \textbf{T}ext-guided \textbf{I}mage \textbf{F}usion (MTIF) in this section. Specifically, Sec.~\ref{ssec:overview} provides an overview of the framework. Sec.~\ref{ssec:data augmentation} presents the semantic-driven visual enrichment module, which enhances the information density and diversity of training data to facilitate more effective multi-modal representation learning. Sec.~\ref{ssec:text-guided fusion} introduces the multi-grained text-guided fusion mechanism, which exploits hierarchical semantic information from textual descriptions to guide fine-grained visual feature fusion, enhancing texture information and semantic content in the final output. Finally, Sec.~\ref{ssec:multi-grained loss} elaborates the multi-grained loss, which is designed to supervise the fusion process across semantic levels, aiming to bridge the semantic gap while preserving structural consistency and texture fidelity.

\subsection{Overview}\label{ssec:overview}
As shown in Fig.~\ref{fig:model}, given an input image pair $\{I_1, I_2\}$, MTIF first applies a semantic-driven visual enrichment module to increase the information density and diversity of the training data, resulting in enriched images $\{I_{ve_1}^n, I_{ve_2}^n\}_{n=1}^N$, where $N$ denotes the number of enriched variants for each input. Subsequently, to obtain deep semantic priors, large language models (LLMs) generate multi-grained textual descriptions that capture detail-level, structure-level, and semantic-level information. These descriptions are encoded by a textual encoder into textual features, denoted as $\{F_t^l\}_{l=1}^L$, where $L$ represents the number of granularity levels. Simultaneously, a visual encoder extracts multi-grained visual features $\{F_{v_1}^l, F_{v_2}^l\}_{l=1}^L$ from enriched image pair $\{I_{ve_1}^n, I_{ve_2}^n\}$. At each granularity, visual features are modulated by the corresponding textual features via the text-guided visual modulation (TGVM) module, yielding text-guided visual features $\{F_{tv_1}^l, F_{tv_2}^l\}_{l=1}^L$. These features are subsequently aggregated to form a fusion feature map $F_f$, which is decoded by a reconstruction module to generate the final fusion image $I_{\text{out}}$.

\begin{figure}[t]
\centering
\includegraphics[width=\linewidth]{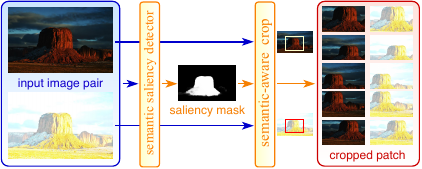}
\vspace{-6mm}
\caption{Pipeline of the Semantic-Driven Visual Enrichment module. First, a semantic saliency detector identifies semantically important regions of the input images. Second, a semantic-aware cropping strategy crops a central region with the highest saliency and four surrounding peripheral patches. This module increases the information density and diversity of the training data and facilitates the learning of multi-grained textual guidance.}
\vspace{+1mm}
\label{fig:crop}
\end{figure}

\subsection{Semantic-Driven Visual Enrichment}\label{ssec:data augmentation}
To enrich the information content during training, we propose a semantic-driven visual enrichment module based on saliency detection~\cite{xie2022pyramid}. Data augmentation techniques adopted by prior methods~\cite{krizhevsky2012imagenet}, such as random cropping, disregard the semantic structure of the image, which may result in omitted targets or semantically incoherent regions. This leads to a lower density of informative content, thereby impairing the potential to learn from semantically significant regions and limiting the effectiveness of multi-grained textual guidance mechanisms. To address this issue, as shown in Fig.~\ref{fig:crop}, we first apply a saliency detection algorithm to process the input image pair $\{I_1, I_2\}$, identifying regions with high semantic importance. Subsequently, we design a center-periphery partitioning strategy, in which the region with the highest saliency response is selected as a central region and four peripheral regions are extended outward from it. The process is formulated as:
\begin{equation}
\{I_{ve_1}^n, I_{ve_2}^n\}_{n=1}^N = \mathcal{VE}(I_1, I_2),
\end{equation}
\noindent
where $\mathcal{VE}(\cdot)$ denotes the visual enrichment operation. This approach not only increases the diversity of training data, but also significantly improves the effective information density of training samples, facilitating the full utilization of multi-grained textual guidance mechanisms.

\subsection{Multi-Grained Text-Guided Fusion}\label{ssec:text-guided fusion}
With the rapid advancement of large language models, multi-modal large language models (MLLM) have exhibited impressive capabilities in understanding visual and textual content and generating appropriate responses. The deep semantic priors embedded in textual descriptions provide valuable guidance for visual feature fusion~\cite{cheng2025textfusion}. However, existing methods typically rely on a coarse-grained textual description, which lacks the semantic depth and diversity. To address this limitation, we propose a multi-grained text-guided fusion mechanism. This design enhances the model’s ability to comprehensively capture fine details, structural layouts, and high-level semantic content of salient regions, thereby improving detailed information and semantic consistency in the fusion image.

Specifically, given a pair of images $\{I_{ve_1}, I_{ve_2}\}$, we first utilize a state-of-the-art MLLM (\textit{i.e.}, GPT-4o~\cite{blog2024hello}) to generate multi-grained semantic descriptions that capture fine-grained details, structural layouts, and high-level semantic concepts of the image pair, denoted as $\{T_{1}^l, T_{2}^l\}_{l=1}^L$. These multi-grained descriptions are subsequently fed into a textual encoder (\textit{i.e.}, BLIP2~\cite{li2023blip}) to extract textual features $\{F_{t}^l\}_{l=1}^L$. This process is defined as follows:
\begin{equation}
\{F_t^l\}_{l=1}^L = \mathcal{E}_t\big( \mathcal{G}(I_1, I_2) \big),
\end{equation}
\noindent
where $\mathcal{G}(\cdot)$ refers to the MLLM and $\mathcal{E}_t(\cdot)$ denotes the textual encoder, respectively. Next, we employ a visual encoder to extract multi-grained visual features $\{F_{v_1}^l, F_{v_2}^l\}_{l=1}^L$. The visual encoder is built upon restormer blocks~\cite{zamir2022restormer} and convolutional neural network (CNN) blocks~\cite{he2016deep}, which efficiently extract both global context and local details from the input images~\cite{Zhao_2024_ICML}. The multi-grained visual features are extracted from different stages of the visual encoder:
\begin{equation}
\{F_{v_1}^l, F_{v_2}^l\}_{l=1}^L = \mathcal{E}_v(I_{ve_1}, I_{ve_2}),
\end{equation} 
\noindent
where $\mathcal{E}_v(\cdot)$ denotes the visual encoder. Subsequently, to enable textual guidance in the modulation of visual features, we introduce a text-guided visual modulation (TGVM) module at each granularity, which modulates visual features by the corresponding textual features to generate text-guided visual features $\{F_{tv_1}^l, F_{tv_2}^l\}_{l=1}^L$. TGVM is implemented via a cross-attention mechanism, where the textual features serve as queries and the visual features act as keys and values. This design enables the textual semantics to guide the visual features, thereby enhancing semantic consistency and representational expressiveness. The process is formulated as:
\begin{align}
F_{tv_1}^l &= \mathcal{TV}(F_{v_1}^l, F_t^l), \nonumber \\
F_{tv_2}^l &= \mathcal{TV}(F_{v_2}^l, F_t^l),
\end{align}
\noindent
where $\mathcal{TV}(\cdot)$ denotes the text-guided visual modulation module. Finally, the text-guided visual features at the last stage $\{F_{tv_1}^L, F_{tv_2}^L\}$ are subsequently processed through $\mathcal{M}(\cdot)$ to obtain the fusion feature map $F_f$, which is then processed by the feature decoder $\mathcal{D}(\cdot)$ to generate the final fusion image $I_{out}$. This process is formulated as:
\begin{equation}
I_{out} = \mathcal{D}\big( \mathcal{M}(F_{tv_1}^L, F_{tv_2}^L) \big),
\end{equation}
\noindent
where $\mathcal{M}(\cdot)$ denotes the feature fusion operation implemented by concatenation along the channel dimension, and $\mathcal{D}(\cdot)$ denotes the decoder.

\begin{figure*}[ht]
    \centering
    \vspace{-8mm}
    \includegraphics[width=\linewidth]
    {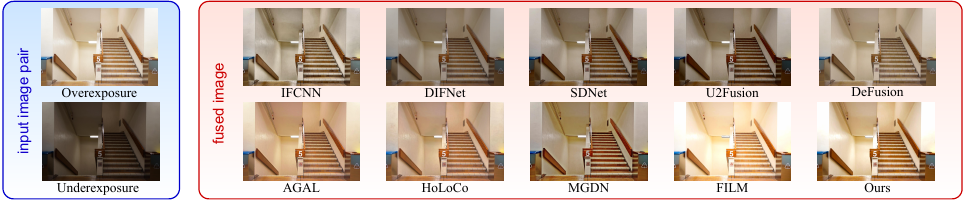}
    \vspace{-6mm}
    \caption{Visualization comparison of fusion results on the SICE~\cite{cai2018learning} dataset for the multi-exposure image fusion task.}
    \label{fig:Qualitative Comparison on SICE}
\end{figure*}

\begin{figure*}[ht]
    \centering
    \includegraphics[width=\linewidth]{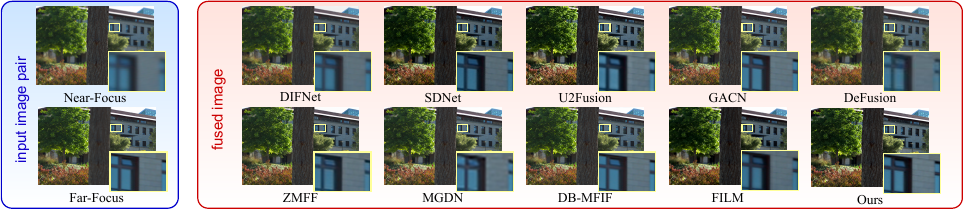}
    \vspace{-7mm}
    \caption{Visualization comparison of fusion outputs on the RealMFF~\cite{zhang2020real} dataset for the multi-focus image fusion task.}
    \vspace{-2mm}
    \label{fig:Qualitative Comparison on RealMFF}
\end{figure*}

\subsection{Multi-Grained Loss for Text-Guided Fusion}\label{ssec:multi-grained loss}
Due to the inherent gap between textual and visual modalities, textual guidance may introduce inconsistencies when modulating visual features. This misalignment hinders the full utilization of information across two modalities. To address this, we propose a multi-grained loss function that enforces hierarchical supervision to progressively bridge the semantic gap at different granularities, thereby enhancing the model’s capacity to jointly capture fine-grained textures and structural semantics.

The loss function of the proposed method $\mathcal{L}_{total}$ consists of two components: (i) the feature reconstruction loss $\mathcal{L}_{feat}$, which performs multi-grained supervision by decoding intermediate multi-modal features into image space and enforcing their consistency with the inputs; and (ii) the base loss $\mathcal{L}_{base}$, which measures the discrepancy between the final fusion image and the input image pair. The overall loss is formulated as:
\begin{equation}
\mathcal{L}_{total} = \mathcal{L}_{feat} + \mathcal{L}_{base}.
\end{equation}

The feature reconstruction loss $\mathcal{L}_{feat}$ is designed to supervise the intermediate text-guided visual features, aiming to bridge the semantic gap between visual and textual modalities and enable more effective textual modulation of visual features. Specifically, for the detail-level features $\{F_{tv_1}^1, F_{tv_2}^1\}$, we decode them into an intermediate fusion image $\hat{I}_{f_1}$. To enforce detail-level consistency with the input images $\{I_{ve_1}, I_{ve_2}\}$, we introduce a gradient loss:
\begin{equation}
\mathcal{L}_{feat\_grad} = \mathcal{L}_{grad}(\hat{I}_{f_1}, \max(I_{ve_1}, I_{ve_2})),
\end{equation}
where $\mathcal{L}_{grad}(\cdot)$ is defined as:
\begin{equation} \label{eq:grad_loss}
\mathcal{L}_{grad}(A, B) = \frac{1}{HW} \left\| |\nabla A| - |\nabla B| \right\|_1,
\end{equation}
with $A$ and $B$ representing the reconstructed and reference images, respectively, $H$ and $W$ denoting the height and width of the image, $\nabla$ representing the Sobel gradient operator, and $\| \cdot \|_1$ denoting the element-wise L1 norm.

For the structure-level text-guided visual features $\{F_{tv_1}^2, F_{tv_2}^2\}$, we decode them into an intermediate reconstruction $\hat{I}_{f_2}$ and compute the structural similarity loss with respect to the input images $\{I_{ve_1}, I_{ve_2}\}$:
\begin{equation}
\mathcal{L}_{feat\_SSIM} = 2 - \text{SSIM}(\hat{I}_{f_2}, I_{ve_1}) - \text{SSIM}(\hat{I}_{f_2}, I_{ve_2})
\end{equation}

The final feature reconstruction loss is a weighted combination of the above components:
\begin{equation}
\mathcal{L}_{feat} = \alpha_1 \mathcal{L}_{feat\_grad} + \alpha_2 \mathcal{L}_{feat\_SSIM},
\end{equation}
\noindent
where $\alpha_{\text{1}}$, $\alpha_{\text{2}}$ are weighting coefficients that balance the contribution of each term.

To supervise the quality of the final fusion image ${I}_{out}$, we introduce the base loss $\mathcal{L}_{base}$, which focuses on preserving both pixel-wise intensity and texture details. It consists of two components: the pixel loss $\mathcal{L}_{base\_pixel}$ and the gradient loss $\mathcal{L}_{base\_grad}$. The pixel loss is defined as:
\begin{equation}
\mathcal{L}_{base\_pixel} = \frac{1}{HW} \left\| I_{out} - \max(I_{ve_1}, I_{ve_2}) \right\|_1.
\end{equation}
The gradient loss is defined as:
\begin{equation}
\mathcal{L}_{base\_grad} = \mathcal{L}_{grad}(I_{out}, \max(I_{ve_1}, I_{ve_2})),
\end{equation}
where $\mathcal{L}_{grad}(\cdot)$ is defined in Eq.~\ref{eq:grad_loss}. The final base loss is computed as a weighted sum of the two components:
\begin{equation}
\mathcal{L}_{base} = \beta_1 \mathcal{L}_{base\_pixel} + \beta_2 \mathcal{L}_{base\_grad},
\end{equation}
where $\beta_1$ and $\beta_2$ are hyper-parameters that balance the contributions of the two terms.
\section{Experiment}
\subsection{Evaluation Settings}
\textbf{Tasks.} We evaluate our method, \textbf{M}ulti-grained \textbf{T}ext-guided \textbf{I}mage \textbf{F}usion (MTIF), on two representative tasks: multi-exposure image fusion (MEF) and multi-focus image fusion (MFF). MEF aims to synthesize images captured at different exposure levels into a single image that effectively retains details in both underexposed and overexposed inputs~\cite{xu2022multi}. MFF seeks to combine images taken under different focal settings to produce an all-in-focus image with sharp content across the entire scene~\cite{zhang2021deep}.

\noindent\textbf{Datasets.} For the MEF task, we use the SICE~\cite{cai2018learning} dataset comprising image pairs captured under varying exposures across diverse natural and urban environments. A subset is used for training and the remainder for testing. Additional validation is conducted on the MEFB~\cite{zhang2021benchmarking} dataset. For the MFF task, we adopt the RealMFF~\cite{zhang2020real} dataset, which consists of real-world image pairs captured with different focal planes, split into training and testing sets. To further assess the generalization capability of our method, we also evaluate on the Lytro~\cite{nejati2015multi} dataset, containing multi-focus image pairs with diverse depth-of-field. Our experimental setup is consistent with prior works~\cite{Zhao_2024_ICML} to ensure a fair comparison.

\noindent\textbf{Evaluation Metrics.}
We adopt six widely used image fusion metrics following the setups in~\cite{Zhao_2024_ICML}. Entropy (EN) measures the amount of information preserved in the fusion image. Standard deviation (SD) reflects the global contrast. Spatial frequency (SF) captures textural details based on frequency variations, while average gradient (AG) quantifies overall image sharpness. Visual information fidelity (VIF) measures the alignment between the fusion image and human visual perception. Qabf evaluates the effectiveness of edge information transfer from the source images to the fusion image~\cite{ma2019infrared}. 

\noindent\textbf{Implementation Details.}  
For the MEF task, the model is trained for 100 epochs using the Adam optimizer. The learning rate is initialized to 8e-5, which is gradually decayed during training. The hyper-parameters $\{\alpha_{1}, \alpha_{2}, \beta_{1}, \beta_{2}\}$ are set to $\{10, 1, 1, 100\}$. The number of granularity levels $L$ is set to 3, and the number of enriched variants for each input $N$ is 5. All experiments are conducted on an NVIDIA A800 GPU. For the MFF task, the model is trained for 45 epochs with the same optimization settings. The hyper-parameters $\{\alpha_{1}, \alpha_{2}, \beta_{1}, \beta_{2}\}$ are adjusted to $\{10, 1, 1, 300\}$. Other training settings remain consistent with those used in the MEF task.

\noindent\textbf{Compared Methods.} 
We comprehensively evaluate MTIF on both MEF and MFF tasks. For the MEF task, we compare with nine representative methods: IFCNN~\cite{zhang2020ifcnn}, DIFNet~\cite{jung2020unsupervised}, SDNet~\cite{zhang2021sdnet}, U2Fusion~\cite{xu2020u2fusion}, DeFusion~\cite{liang2022fusion}, AGAL~\cite{liu2022attention}, HoLoCo~\cite{liu2023holoco}, MGDN~\cite{guan2023mutual}, and FILM~\cite{Zhao_2024_ICML}. For the MFF task, we compare MTIF with nine recent methods: DIFNet~\cite{jung2020unsupervised}, SDNet~\cite{zhang2021sdnet}, U2Fusion~\cite{xu2020u2fusion}, 
GACN~\cite{ma2022end}, DeFusion~\cite{liang2022fusion}, ZMFF~\cite{hu2023zmff}, MGDN~\cite{guan2023mutual}, DB-MFIF~\cite{zhang2024exploit}, and FILM~\cite{Zhao_2024_ICML}. These methods encompass a diverse range of network architectures and fusion strategies, ensuring a comprehensive and fair comparison.

\subsection{Multi-Exposure Image Fusion}
\textbf{Qualitative Comparison.}
Fig.~\ref{fig:Qualitative Comparison on SICE} presents the fusion outputs of our method and state-of-the-art methods on the MEF dataset. As shown in Fig.~\ref{fig:Qualitative Comparison on SICE}, the fusion outputs from most existing methods (\eg, IFCNN~\cite{zhang2020ifcnn}, SDNet~\cite{zhang2021sdnet}, U2Fusion~\cite{xu2020u2fusion}) exhibit insufficient brightness. In contrast, FILM~\cite{Zhao_2024_ICML} generates overly bright outputs with noticeable overexposure in the wall region. Although AGAL~\cite{liu2022attention} and HoLoCo~\cite{liu2023holoco} preserve appropriate luminance, their results suffer from low contrast and lack sufficient details. In comparison, our method demonstrates superior performance in maintaining brightness and contrast, while effectively preserving details, leading to more natural and visually compelling fusion outputs. This demonstrates that the multi-grained textual guidance in our method integrates hierarchical semantic priors to guide visual feature fusion at corresponding levels, with the additional integration of the hierarchical supervision loss and the saliency-driven visual enrichment module, which together enable better preservation of fine details and structural consistency in the fusion outputs. Additional qualitative results are provided in Sec.~\ref{sec:suppl_mef}.

\begin{table}[ht]
        \centering
        \renewcommand\arraystretch{0.96}
        \caption{Quantitative results of multi-exposure image fusion on the SICE~\cite{cai2018learning} dataset. The \textbf{bold} markers represent the best values.}
        \vspace{-2mm}
        \label{tab:MEF_SICE}%
        \setlength{\tabcolsep}{10pt}
        \resizebox{1.0\linewidth}{!}{
            \begin{tabular}{crrrrrrcrrrrrr}
                \toprule
                \multicolumn{7}{c}{\textbf{SICE Multi-exposure Image Fusion Dataset}}\\
                & \multicolumn{1}{c}{EN} & \multicolumn{1}{c}{SD} & \multicolumn{1}{c}{SF} & \multicolumn{1}{c}{AG} & \multicolumn{1}{c}{VIF} & \multicolumn{1}{c}{Qabf}\\
                \midrule
                IFCNN~\cite{zhang2020ifcnn} & 6.67  & 39.43 & 16.93 & 4.59  & 0.73  & 0.71\\
                DIFNet~\cite{jung2020unsupervised} & 6.56  & 35.76 & 11.86 & 3.09  & 0.46  & 0.50\\
                SDNet~\cite{zhang2021sdnet} & 6.47  & 38.25 & 19.34 & 4.80  & 0.48  & 0.45\\
                U2Fusion~\cite{xu2020u2fusion} & 6.43  & 34.77 & 10.71 & 3.17  & 0.48  & 0.57\\
                DeFusion~\cite{liang2022fusion} & 6.87  & 44.73 & 14.28 & 4.04  & 0.87  & 0.57\\
                AGAL~\cite{liu2022attention}  & 7.06 & 46.03 & 16.64 & 4.91 & 0.72  & 0.53\\
                HoLoCo~\cite{liu2023holoco} & 7.04  & 42.73 & 9.33  & 3.47  & 0.74  & 0.37\\
                MGDN~\cite{guan2023mutual}  & 6.94  & 43.69 & 15.04 & 4.59  & 0.88 & 0.64\\
                FILM~\cite{Zhao_2024_ICML}  & 7.07 & 54.21 & \textbf{19.42} & 5.15 & 1.05 & 0.79\\
                \textbf{Ours}  & \textbf{7.28} & \textbf{60.41} & 18.78 & \textbf{5.40} & \textbf{1.46} & \textbf{0.79}\\
                \bottomrule
        \end{tabular}}
        \vspace{-2mm}
    \end{table}%

\begin{table}[ht]
        \centering
        \renewcommand\arraystretch{0.96}
        \vspace{-1mm}
        \caption{Quantitative results of multi-exposure image fusion on the MEFB~\cite{zhang2021benchmarking} dataset. The \textbf{bold} markers represent the best values.}
        \vspace{-2mm}
        \label{tab:MEF_MEFB}%
        \setlength{\tabcolsep}{10pt}
        \resizebox{1.0\linewidth}{!}{
            \begin{tabular}{crrrrrrcrrrrrr}
                \toprule
                 \multicolumn{7}{c}{\textbf{MEFB Multi-exposure Image Fusion Dataset}} \\
                & \multicolumn{1}{c}{EN} & \multicolumn{1}{c}{SD} & \multicolumn{1}{c}{SF} & \multicolumn{1}{c}{AG} & \multicolumn{1}{c}{VIF} & \multicolumn{1}{c}{Qabf} \\
                \midrule
                IFCNN~\cite{zhang2020ifcnn} & 6.99  & 52.49 & 18.16 & 5.34  & 0.71  & 0.69 \\
                DIFNet~\cite{jung2020unsupervised} & 6.99  & 50.23 & 11.79 & 3.47  & 0.51  & 0.53 \\
                SDNet~\cite{zhang2021sdnet} & 6.59  & 51.77 & 20.53 & 5.27  & 0.55  & 0.42 \\
                U2Fusion~\cite{xu2020u2fusion} & 6.67  & 46.73 & 12.54 & 3.82  & 0.51  & 0.56 \\
                DeFusion~\cite{liang2022fusion} & 7.10  & 56.46 & 14.86 & 4.48  & 0.70  & 0.59 \\
                AGAL~\cite{liu2022attention}  & 7.14  & 60.63 & 17.77 & 5.33  & 0.79  & 0.65 \\
                HoLoCo~\cite{liu2023holoco} & 7.20  & 53.88 & 12.80 & 4.34  & 0.73  & 0.54 \\
                MGDN~\cite{guan2023mutual}  & 7.25 & 55.97 & 18.09 & 5.76 & 0.96 & 0.65 \\
                FILM~\cite{Zhao_2024_ICML}  & 7.31 & 69.02 & 20.98 & 6.15 & 0.98 & \textbf{0.77} \\
                \textbf{Ours}  & \textbf{7.36} & \textbf{69.71} & \textbf{22.09} & \textbf{6.35} & \textbf{1.35} & 0.75 \\
                \bottomrule
        \end{tabular}}
        \vspace{-2mm}
    \end{table}%

    \begin{table}[t]
        \centering
        \renewcommand\arraystretch{0.96}
        \caption{Quantitative results of multi-focus image fusion on the RealMFF~\cite{zhang2020real} dataset. The \textbf{bold} markers represent the best values.}
        \vspace{-2mm}
        \label{tab:MFF_RealMFF}%
        \setlength{\tabcolsep}{10pt}
        \resizebox{\linewidth}{!}{
            \begin{tabular}{crrrrrrcrrrrrr}
                \toprule
                \multicolumn{7}{c}{\textbf{RealMFF Multi-focus Image Fusion Dataset}}\\
                & \multicolumn{1}{c}{EN} & \multicolumn{1}{c}{SD} & \multicolumn{1}{c}{SF} & \multicolumn{1}{c}{AG} & \multicolumn{1}{c}{VIF} & \multicolumn{1}{c}{Qabf}\\
                \midrule
                DIFNet~\cite{jung2020unsupervised} & 7.01  & 51.17 & 10.78 & 3.96  & 0.89  & 0.69 \\
                SDNet~\cite{zhang2021sdnet} & 6.95  & 50.96 & 15.22 & 5.02  & 0.93  & 0.73\\
                U2Fusion~\cite{xu2020u2fusion} & 6.77  & 48.49 & 14.07 & 5.09  & 0.95  & 0.70 \\
                GACN~\cite{ma2022end} & 6.99  & 51.59 & 14.93 & 5.00  & 1.53  & 0.76\\
                DeFusion~\cite{liang2022fusion} & 7.09  & 54.42 & 11.24 & 4.08  & 0.98  & 0.69\\
                ZMFF~\cite{hu2023zmff}  & 6.99  & 51.15 & 13.93 & 4.95  & 0.94  & 0.70\\
                MGDN~\cite{guan2023mutual}  & 7.09 & 54.24 & 15.15 & 5.24 & 1.07 & 0.75\\
                DB-MFIF~\cite{zhang2024exploit}   & 7.02  & 51.85 & 15.39 & 5.16  & 1.50  & 0.75\\
                FILM~\cite{Zhao_2024_ICML}  & 7.11 & 54.93 & 15.62 & 5.43 & 1.10 & 0.76\\
                \textbf{Ours}  & \textbf{7.14} & \textbf{56.03} & \textbf{15.78} & \textbf{5.43} & \textbf{1.56} & \textbf{0.77}\\
                \bottomrule
        \end{tabular}}
        \vspace{-2mm}
    \end{table}%
    \begin{table}[ht]
        \centering
        \renewcommand\arraystretch{0.96}
        \caption{Quantitative results of multi-focus image fusion on the Lytro~\cite{nejati2015multi} dataset. The \textbf{bold} markers represent the best values.}
        \vspace{-2mm}
        \label{tab:MFF_Lytro}%
        \setlength{\tabcolsep}{10pt}
        \resizebox{\linewidth}{!}{
            \begin{tabular}{crrrrrrcrrrrrr}
                \toprule
                \multicolumn{7}{c}{\textbf{Lytro Multi-focus Image Fusion Dataset}} \\
                & \multicolumn{1}{c}{EN} & \multicolumn{1}{c}{SD} & \multicolumn{1}{c}{SF} & \multicolumn{1}{c}{AG} & \multicolumn{1}{c}{VIF} & \multicolumn{1}{c}{Qabf} \\
                \midrule
                DIFNet~\cite{jung2020unsupervised} & 7.43  & 52.52 & 11.47 & 4.30  & 0.73  & 0.54 \\
                SDNet~\cite{zhang2021sdnet} & 7.47  & 55.25 & 16.88 & 5.84  & 0.84  & 0.69 \\
                U2Fusion~\cite{xu2020u2fusion} & 7.30  & 51.95 & 14.83 & 5.60  & 0.83  & 0.65 \\
                GACN~\cite{ma2022end} & 7.53  & 57.47 & 19.36 & 6.79  & \textbf{1.37}  & 0.75 \\
                DeFusion~\cite{liang2022fusion} & 7.52  & 56.65 & 11.55 & 4.35  & 0.80  & 0.55 \\
                ZMFF~\cite{hu2023zmff}  & 7.53  & 56.96 & 18.84 & 6.76  & 0.93  & 0.69 \\
                MGDN~\cite{guan2023mutual}  & 7.54 & 57.50 & 18.81 & 6.67  & 0.93  & 0.74 \\
                DB-MFIF~\cite{zhang2024exploit}   & 7.54  & 57.71 & 19.56 & 6.86  & 1.35 & 0.74 \\
                FILM~\cite{Zhao_2024_ICML}  & 7.56 & 59.15 & 19.57 & 6.97 & 0.98 & 0.74 \\
                \textbf{Ours}  & \textbf{7.59} & \textbf{60.85} & \textbf{19.85} & \textbf{7.01} & 1.30 & \textbf{0.75} \\
                \bottomrule
        \end{tabular}}
        \vspace{-2mm}
    \end{table}%

\begin{table*}[t]
    \centering
    \renewcommand\arraystretch{0.83}
    \vspace{-4mm}
    \caption{Ablation of key components on multi-exposure and multi-focus image fusion tasks. The \textbf{bold} markers represent the best values.}
    \vspace{-2mm}
    \label{tab:ablation_components}
    \setlength{\tabcolsep}{10pt}
    \small
    \resizebox{1.0\linewidth}{!}{
    \begin{tabular}{l|cccccc|cccccc}
        \toprule
        \multirow{2}{*}{Configuration} 
        & \multicolumn{6}{c|}{\textbf{SICE Multi-exposure Image Fusion Dataset}} 
        & \multicolumn{6}{c}{\textbf{RealMFF Multi-focus Image Fusion Dataset}} \\
        \cmidrule(lr){2-7} \cmidrule(lr){8-13}
        & EN & SD & SF & AG & VIF & Qabf 
        & EN & SD & SF & AG & VIF & Qabf \\
        \midrule
        Exp.~I: ~~w/o TG 
        & 7.15  & 58.26 & 19.18 & 5.40  & 1.42  & 0.75 
        & 7.13  & 55.54 & 15.56 & 5.36  & 1.56  & 0.77 \\
        
        Exp.~II: ~w/o ML 
        & 7.14  & 58.43 & \textbf{19.27} & 5.39  & 1.41  & 0.75 
        & 7.13  & 55.82 & 15.59 & 5.37  & 1.56  & 0.77 \\
        
        Exp.~III: w/o VE 
        & 7.24  & 59.54 & 18.88 & 5.37  & 1.45  & 0.77 
        & 7.14  & 56.01 & 15.68 & 5.41  & \textbf{1.59}  & 0.76 \\

        Exp.~IV: replace VE with RC 
        & 7.13  & 58.69 & 19.21 & \textbf{5.55}  & 1.44  & 0.76 
        & 7.12  & 55.43 & 15.60 & 5.36  & 1.56  & 0.77 \\
        
        \midrule
        \textbf{Ours}
        & \textbf{7.28} & \textbf{60.41} & 18.78 & 5.40 & \textbf{1.46} & \textbf{0.79}
        & \textbf{7.14} & \textbf{56.03} & \textbf{15.78} & \textbf{5.43} & 1.56 & \textbf{0.77} \\
        \bottomrule
    \end{tabular}}
    \vspace{-2mm}
\end{table*}

\noindent\textbf{Quantitative Comparison.}
Tabs.~\ref{tab:MEF_SICE} and~\ref{tab:MEF_MEFB} present the quantitative results of all compared methods on two benchmark datasets. Our method achieves the best performance across five metrics on each dataset, demonstrating strong generalization and robustness. 
On the SICE~\cite{cai2018learning} dataset, the improvements in EN, SD, and AG reflect our ability to enhance global information content, contrast, and sharpness, while the gains in VIF and Qabf demonstrate superior perceptual fidelity and edge preservation. The competitive SF score confirms that our method preserves spatial texture and edge details while minimizing artifacts. On the MEFB~\cite{zhang2021benchmarking} dataset, MTIF consistently outperforms other methods across all metrics, except Qabf, where it remains highly competitive. This demonstrates that our fusion outputs effectively preserve fine details and maintain high perceptual quality, even in the presence of challenging exposure variations. The consistent high scores across diverse metrics underscore the effectiveness of our multi-grained textual guidance and hierarchical supervision strategies.

\subsection{Multi-Focus Image Fusion}
\textbf{Qualitative Comparison.}
Fig.~\ref{fig:Qualitative Comparison on RealMFF} shows the qualitative comparison results of our method against nine state-of-the-art methods on the MFF dataset. As shown in Fig.~\ref{fig:Qualitative Comparison on RealMFF}, the fusion image produced by GACN~\cite{ma2022end} appears overly bright, while DIFNet~\cite{jung2020unsupervised}, DeFusion~\cite{liang2022fusion}, and MGDN~\cite{guan2023mutual} fail to preserve sufficient detail, resulting in blurry textures on the wall surface. SDNet~\cite{zhang2021sdnet}, U2Fusion~\cite{xu2020u2fusion}, ZMFF~\cite{hu2023zmff}, DB-MFIF~\cite{zhang2024exploit}, and FILM~\cite{Zhao_2024_ICML} suffer from noticeable artifacts, particularly around the window edges. In contrast, our method achieves superior brightness and detail preservation, particularly on surface areas like walls, while effectively avoiding artifacts, leading to cleaner and more visually faithful fusion outputs. The superior performance is driven by the multi-grained textual guidance mechanism, which injects hierarchical semantic priors to modulate visual features. The hierarchical supervision loss bridges the semantic gap between modalities, enhancing alignment of textual and visual features. Additionally, the saliency-driven visual enrichment module increases the density of informative content, improving the model's focus on semantically critical regions. Together, these components enable high-quality fusion results with natural appearance, rich details, and minimal artifacts. Additional qualitative results are provided in Sec.~\ref{sec:suppl_mff}.

\begin{figure}[t]
    \centering
    \includegraphics[width=\linewidth]{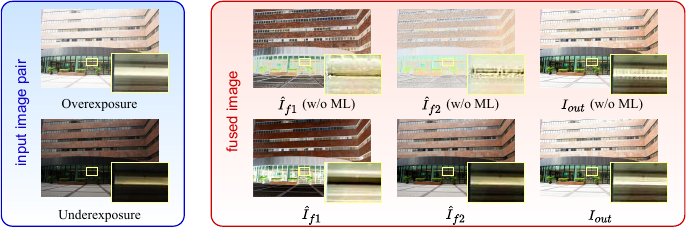}
    \vspace{-6mm}
    \caption{Visualization of the impact of multi-grained loss on fusion outputs at each semantic level.}
    \vspace{-2mm}
    \label{fig:Ablation ML}
\end{figure}

\noindent\textbf{Quantitative Comparison.}
Tabs.~\ref{tab:MFF_RealMFF} and~\ref{tab:MFF_Lytro} present the quantitative results of all compared methods on the MFF task across the RealMFF~\cite{zhang2020real} and Lytro~\cite{nejati2015multi} datasets. On the RealMFF~\cite{zhang2020real} dataset, our method achieves the highest scores across all six metrics, demonstrating superior fine detail preservation and strong alignment with human visual perception. On Lytro~\cite{nejati2015multi}, MTIF outperforms all baselines in EN, SD, SF, AG, and Qabf, demonstrating improved sharpness and contrast. It also performs competitively on VIF, further validating its strong perceptual fidelity. The consistently superior performance across both datasets highlights the robustness and generalization ability of our approach in handling diverse multi-focus scenarios.

\subsection{Ablation Studies}
To validate the effectiveness of our approach, we conduct ablation studies on the SICE~\cite{cai2018learning} and RealMFF~\cite{zhang2020real} datasets. Specifically, we assess the impact of three key components: multi-grained textual guidance (TG), multi-grained loss (ML), and visual information enrichment (VE), by removing each component individually. As shown in Tab.~\ref{tab:ablation_components}, in Exp.~I, we remove TG to evaluate its contribution. In Exp.~II, we use the base loss (denoted as $\mathcal{L}_{base}$) to evaluate the contribution of ML. In Exp.~III, we remove the VE module to assess its impact on the effectiveness of the multi-grained textual guidance mechanism in improving visual feature fusion. In Exp.~IV, we replace VE with random crop augmentation (RC) to further assess the effectiveness of enhancing the semantic density and structural diversity of the training data in improving model performance. Overall, the ablation results demonstrate that the semantic information from multi-grained textual descriptions effectively guides the fusion of visual features, while the hierarchical loss and semantic-driven visual enrichment further enhance the model’s ability to align and leverage multi-modal information. Additionally, as shown in Fig.~\ref{fig:Ablation ML}, ablating ML results in accumulated artifacts and degraded details in the bright areas, whereas incorporating ML leads to cleaner outputs with improved perceptual quality at each semantic level. These results demonstrate the superiority of MTIF.

\section{Conclusion}
In this paper, we propose \textbf{M}ulti-Grained \textbf{T}ext-Guided \textbf{I}mage \textbf{F}usion (\textbf{MTIF}), a novel image fusion framework that introduces multi-grained textual guidance to enhance the fusion of multi-exposure and multi-focus image pairs. By leveraging multi-grained textual descriptions, our approach effectively guides image fusion through a hierarchical cross-modal modulation module. Additionally, we design a multi-grained supervision strategy and a saliency-driven enrichment module to further strengthen cross-modal alignment and modulation. Extensive experiments demonstrate the effectiveness of our approach, achieving strong performance compared to state-of-the-art methods. This work provides new insights for integrating vision-language models into low-level vision tasks.

\section*{Acknowledgments}
This work was supported in part by the National Natural Science Foundation of China under Grants No. U25A20531 and U22A2096.

{
    \small
    \bibliographystyle{ieeenat_fullname}
    \bibliography{main}

@String(IJCV = {Int. J. Comput. Vis.})

@String(CVPR= {IEEE Conf. Comput. Vis. Pattern Recog.})

@String(ICCV= {Int. Conf. Comput. Vis.})

@String(ECCV= {Eur. Conf. Comput. Vis.})

@String(TIP  = {IEEE Trans. Image Process.})

@String(TMM  = {IEEE Trans. Multimedia})

@String(ACMMM= {ACM Int. Conf. Multimedia})

@String(ICIP = {IEEE Int. Conf. Image Process.})

@String(ICLR = {Int. Conf. Learn. Represent.})

@String(AAAI = {AAAI})

@String(IJCV  = {IJCV})

@String(CVPR  = {CVPR})

@String(ICCV  = {ICCV})

@String(ECCV  = {ECCV})

@String(TIP   = {IEEE TIP})

@String(TCSVT = {IEEE TCSVT})

@String(TMM   =	{IEEE TMM})

@String(ACMMM = {ACM MM})

@String(ICIP  = {ICIP})

@String(ICLR  = {ICLR})

@article{xu2022multi,
  title={Multi-exposure image fusion techniques: A comprehensive review},
  author={Xu, Fang and Liu, Jinghong and Song, Yueming and Sun, Hui and Wang, Xuan},
  journal={Remote Sensing},
  year={2022}
}

@article{song2011probabilistic,
  title={Probabilistic exposure fusion},
  author={Song, Mingli and Tao, Dacheng and Chen, Chun and Bu, Jiajun and Luo, Jiebo and Zhang, Chengqi},
  journal={TIP},
  year={2011}
}

@inproceedings{ma2015multi,
  title={Multi-exposure image fusion: A patch-wise approach},
  author={Ma, Kede and Wang, Zhou},
  booktitle={ICIP},
  year={2015}
}

@article{ma2017robust,
  title={Robust multi-exposure image fusion: A structural patch decomposition approach},
  author={Ma, Kede and Li, Hui and Yong, Hongwei and Wang, Zhou and Meng, Deyu and Zhang, Lei},
  journal={TIP},
  year={2017}
}

@article{ma2017multi,
  title={Multi-exposure image fusion by optimizing a structural similarity index},
  author={Ma, Kede and Duanmu, Zhengfang and Yeganeh, Hojatollah and Wang, Zhou},
  journal={TCI},
  year={2017}
}

@article{li2020fast,
  title={Fast multi-scale structural patch decomposition for multi-exposure image fusion},
  author={Li, Hui and Ma, Kede and Yong, Hongwei and Zhang, Lei},
  journal={TIP},
  year={2020}
}

@article{ulucan2021multi,
  title={Multi-exposure image fusion based on linear embeddings and watershed masking},
  author={Ulucan, Oguzhan and Karakaya, Diclehan and Turkan, Mehmet},
  journal={Signal Processing},
  year={2021}
}

@article{shen2012qoe,
  title={QoE-based multi-exposure fusion in hierarchical multivariate Gaussian CRF},
  author={Shen, Rui and Cheng, Irene and Basu, Anup},
  journal={TIP},
  year={2012}
}

@article{li2013image,
  title={Image fusion with guided filtering},
  author={Li, Shutao and Kang, Xudong and Hu, Jianwen},
  journal={TIP},
  year={2013}
}

@article{shen2014exposure,
  title={Exposure fusion using boosting Laplacian pyramid},
  author={Shen, Jianbing and Zhao, Ying and Yan, Shuicheng and Li, Xuelong},
  journal={TCYB},
  year={2014}
}

@inproceedings{nejati2017fast,
  title={Fast exposure fusion using exposedness function},
  author={Nejati, Mansour and Karimi, Maryam and Soroushmehr, SM Reza and Karimi, Nader and Samavi, Shadrokh and Najarian, Kayvan},
  booktitle={ICIP},
  year={2017}
}

@article{li2017detail,
  title={Detail-enhanced multi-scale exposure fusion},
  author={Li, Zhengguo and Wei, Zhe and Wen, Changyun and Zheng, Jinghong},
  journal={TIP},
  year={2017}
}

@article{wang2019detail,
  title={Detail-enhanced multi-scale exposure fusion in YUV color space},
  author={Wang, Qiantong and Chen, Weihai and Wu, Xingming and Li, Zhengguo},
  journal={TCSVT},
  year={2019}
}

@article{wang2016multi,
  title={Multi-class remote sensing object recognition based on discriminative sparse representation},
  author={Wang, Xin and Shen, Siqiu and Ning, Chen and Huang, Fengchen and Gao, Hongmin},
  journal={Applied Optics},
  year={2016}
}

@article{yang2019multiexposure,
  title={Multiexposure estimation and fusion based on a sparsity exposure dictionary},
  author={Yang, Yong and Wu, Jiahua and Huang, Shuying and Lin, Pan},
  journal={TIM},
  year={2019}
}

@article{martorell2019ghosting,
  title={Ghosting-free DCT based multi-exposure image fusion},
  author={Martorell, Onofre and Sbert, Catalina and Buades, Antoni},
  journal={SPIC},
  year={2019}
}

@inproceedings{ram2017deepfuse,
  title={Deepfuse: A deep unsupervised approach for exposure fusion with extreme exposure image pairs},
  author={Ram Prabhakar, K and Sai Srikar, V and Venkatesh Babu, R},
  booktitle={ICCV},
  year={2017}
}

@article{zhang2020ifcnn,
  title={Ifcnn: A general image fusion framework based on convolutional neural network},
  author={Zhang, Yu and Liu, Yu and Sun, Peng and Yan, Han and Zhao, Xiaolin and Zhang, Li},
  journal={Information Fusion},
  year={2020}
}

@article{xu2020mef,
  title={Mef-gan: Multi-exposure image fusion via generative adversarial networks},
  author={Xu, Han and Ma, Jiayi and Zhang, Xiao-Ping},
  journal={TIP},
  year={2020}
}

@article{liu2023holoco,
  title={Holoco: Holistic and local contrastive learning network for multi-exposure image fusion},
  author={Liu, Jinyuan and Wu, Guanyao and Luan, Junsheng and Jiang, Zhiying and Liu, Risheng and Fan, Xin},
  journal={Information Fusion},
  year={2023}
}

@inproceedings{xu2023unsupervised,
  title={Unsupervised multi-exposure image fusion breaking exposure limits via contrastive learning},
  author={Xu, Han and Haochen, Liang and Ma, Jiayi},
  booktitle={AAAI},
  year={2023}
}

@inproceedings{Zhao_2024_ICML,
    title={Image fusion via vision-language model},
    author={Zixiang Zhao and Lilun Deng and Haowen Bai and Yukun Cui and Zhipeng Zhang 
            and Yulun Zhang and Haotong Qin and Dongdong Chen and Jiangshe Zhang 
            and Peng Wang and Luc Van Gool},
    booktitle={ICML},
    year={2024}
}

@article{zhang2021deep,
  title={Deep learning-based multi-focus image fusion: A survey and a comparative study},
  author={Zhang, Xingchen},
  journal={TPAMI},
  year={2021}
}

@article{petrovic2004gradient,
  title={Gradient-based multiresolution image fusion},
  author={Petrovic, Vladimir S and Xydeas, Costas S},
  journal={TIP},
  year={2004}
}

@article{phamila2013low,
  title={Low complexity multifocus image fusion in discrete cosine transform domain},
  author={Phamila, Asnath Victy and Amutha, R},
  journal={Optica Applicata},
  year={2013}
}

@article{yang2009multifocus,
  title={Multifocus image fusion and restoration with sparse representation},
  author={Yang, Bin and Li, Shutao},
  journal={TIM},
  year={2009}
}

@article{li2001combination,
  title={Combination of images with diverse focuses using the spatial frequency},
  author={Li, Shutao and Kwok, James T and Wang, Yaonan},
  journal={Information Fusion},
  year={2001}
}

@article{bai2015quadtree,
  title={Quadtree-based multi-focus image fusion using a weighted focus-measure},
  author={Bai, Xiangzhi and Zhang, Yu and Zhou, Fugen and Xue, Bindang},
  journal={Information Fusion},
  year={2015}
}

@article{zhang2017boundary,
  title={Boundary finding based multi-focus image fusion through multi-scale morphological focus-measure},
  author={Zhang, Yu and Bai, Xiangzhi and Wang, Tao},
  journal={Information Fusion},
  year={2017}
}

@article{liu2015multi,
  title={Multi-focus image fusion with dense SIFT},
  author={Liu, Yu and Liu, Shuping and Wang, Zengfu},
  journal={Information Fusion},
  year={2015}
}

@article{li2013image_dynamic_scenes,
  title={Image matting for fusion of multi-focus images in dynamic scenes},
  author={Li, Shutao and Kang, Xudong and Hu, Jianwen and Yang, Bin},
  journal={Information Fusion},
  year={2013}
}

@article{liu2020multi,
  title={Multi-focus image fusion: A survey of the state of the art},
  author={Liu, Yu and Wang, Lei and Cheng, Juan and Li, Chang and Chen, Xun},
  journal={Information Fusion},
  year={2020}
}

@article{liu2017multi,
  title={Multi-focus image fusion with a deep convolutional neural network},
  author={Liu, Yu and Chen, Xun and Peng, Hu and Wang, Zengfu},
  journal={Information Fusion},
  year={2017}
}

@article{zhang2021mff,
  title={Mff-gan: An unsupervised generative adversarial network with adaptive and gradient joint constraints for multi-focus image fusion},
  author={Zhang, Hao and Le, Zhuliang and Shao, Zhenfeng and Xu, Han and Ma, Jiayi},
  journal={Information Fusion},
  year={2021}
}

@article{hu2023zmff,
  title={Zmff: Zero-shot multi-focus image fusion},
  author={Hu, Xingyu and Jiang, Junjun and Liu, Xianming and Ma, Jiayi},
  journal={Information Fusion},
  year={2023}
}

@inproceedings{radford2021learning,
  title={Learning transferable visual models from natural language supervision},
  author={Radford, Alec and Kim, Jong Wook and Hallacy, Chris and Ramesh, Aditya and Goh, Gabriel and Agarwal, Sandhini and Sastry, Girish and Askell, Amanda and Mishkin, Pamela and Clark, Jack and Krueger, Gretchen and Sutskever, Ilya},
  booktitle={ICML},
  year={2021}
}

@inproceedings{li2022blip,
  title={Blip: Bootstrapping language-image pre-training for unified vision-language understanding and generation},
  author={Li, Junnan and Li, Dongxu and Xiong, Caiming and Hoi, Steven},
  booktitle={ICML},
  year={2022}
}

@inproceedings{li2023blip,
  title={Blip-2: Bootstrapping language-image pre-training with frozen image encoders and large language models},
  author={Li, Junnan and Li, Dongxu and Savarese, Silvio and Hoi, Steven},
  booktitle={ICML},
  year={2023}
}

@inproceedings{liu2023visual,
  title={Visual instruction tuning},
  author={Liu, Haotian and Li, Chunyuan and Wu, Qingyang and Lee, Yong Jae},
  booktitle={NeurIPS},
  year={2023}
}

@inproceedings{wu2024visionllm,
  title={Visionllm v2: An end-to-end generalist multimodal large language model for hundreds of vision-language tasks},
  author={Wu, Jiannan and Zhong, Muyan and Xing, Sen and Lai, Zeqiang and Liu, Zhaoyang and Chen, Zhe and Wang, Wenhai and Zhu, Xizhou and Lu, Lewei and Lu, Tong and Luo, Ping and Qiao, Yu and Dai, Jifeng},
  booktitle={NeurIPS},
  year={2024}
}

@inproceedings{wu2024general,
  title={General object foundation model for images and videos at scale},
  author={Wu, Junfeng and Jiang, Yi and Liu, Qihao and Yuan, Zehuan and Bai, Xiang and Bai, Song},
  booktitle={CVPR},
  year={2024}
}

@inproceedings{wang2024llm,
  title={Llm-seg: Bridging image segmentation and large language model reasoning},
  author={Wang, Junchi and Ke, Lei},
  booktitle={CVPR},
  year={2024}
}

@inproceedings{agenticir,
      title={An intelligent agentic system for complex image restoration problems1},
      author={Kaiwen Zhu and Jinjin Gu and Zhiyuan You and Yu Qiao and Chao Dong},
      booktitle={ICLR},
      year={2025}
}

@inproceedings{sun2024coser,
  title={Coser: Bridging image and language for cognitive super-resolution},
  author={Sun, Haoze and Li, Wenbo and Liu, Jianzhuang and Chen, Haoyu and Pei, Renjing and Zou, Xueyi and Yan, Youliang and Yang, Yujiu},
  booktitle={CVPR},
  year={2024}
}

@article{wang2025multi,
  title={Multi-text guidance is important: Multi-modality image fusion via large generative vision-language model},
  author={Wang, Zeyu and Zhao, Libo and Zhang, Jizheng and Song, Rui and Song, Haiyu and Meng, Jiana and Wang, Shidong},
  journal={IJCV},
  year={2025}
}

@article{cheng2025textfusion,
  title={Textfusion: Unveiling the power of textual semantics for controllable image fusion},
  author={Cheng, Chunyang and Xu, Tianyang and Wu, Xiao-Jun and Li, Hui and Li, Xi and Tang, Zhangyong and Kittler, Josef},
  journal={Information Fusion},
  year={2025}
}

@inproceedings{yi2024text,
  title={Text-if: Leveraging semantic text guidance for degradation-aware and interactive image fusion},
  author={Yi, Xunpeng and Xu, Han and Zhang, Hao and Tang, Linfeng and Ma, Jiayi},
  booktitle={CVPR},
  year={2024}
}

@article{li2023text,
  title={From text to pixels: A context-aware semantic synergy solution for infrared and visible image fusion},
  author={Li, Xingyuan and Zou, Yang and Liu, Jinyuan and Jiang, Zhiying and Ma, Long and Fan, Xin and Liu, Risheng},
  journal={arXiv preprint arXiv:2401.00421},
  year={2023}
}

@article{blog2024hello,
  title={Introducing ChatGPT},
  author={Blog, O},
  journal={Internet: https://openai.com/index/hello-gpt-4o/},
  year={2024}
}

@inproceedings{zamir2022restormer,
  title={Restormer: Efficient transformer for high-resolution image restoration},
  author={Zamir, Syed Waqas and Arora, Aditya and Khan, Salman and Hayat, Munawar and Khan, Fahad Shahbaz and Yang, Ming-Hsuan},
  booktitle={CVPR},
  year={2022}
}

@article{cai2018learning,
  title={Learning a deep single image contrast enhancer from multi-exposure images},
  author={Cai, Jianrui and Gu, Shuhang and Zhang, Lei},
  journal={TIP},
  year={2018}
}

@article{zhang2021benchmarking,
  title={Benchmarking and comparing multi-exposure image fusion algorithms},
  author={Zhang, Xingchen},
  journal={Information Fusion},
  year={2021}
}

@article{zhang2020real,
  title={Real-mff: A large realistic multi-focus image dataset with ground truth},
  author={Zhang, Juncheng and Liao, Qingmin and Liu, Shaojun and Ma, Haoyu and Yang, Wenming and Xue, Jing-Hao},
  journal={PRL},
  year={2020}
}

@article{nejati2015multi,
  title={Multi-focus image fusion using dictionary-based sparse representation},
  author={Nejati, Mansour and Samavi, Shadrokh and Shirani, Shahram},
  journal={Information Fusion},
  year={2015}
}

@article{ma2019infrared,
  title={Infrared and visible image fusion methods and applications: A survey},
  author={Ma, Jiayi and Ma, Yong and Li, Chang},
  journal={Information Fusion},
  year={2019}
}

@article{jung2020unsupervised,
  title={Unsupervised deep image fusion with structure tensor representations},
  author={Jung, Hyungjoo and Kim, Youngjung and Jang, Hyunsung and Ha, Namkoo and Sohn, Kwanghoon},
  journal={TIP},
  year={2020}
}

@article{zhang2021sdnet,
  title={Sdnet: A versatile squeeze-and-decomposition network for real-time image fusion},
  author={Zhang, Hao and Ma, Jiayi},
  journal={IJCV},
  year={2021}
}

@article{xu2020u2fusion,
  title={U2fusion: A unified unsupervised image fusion network},
  author={Xu, Han and Ma, Jiayi and Jiang, Junjun and Guo, Xiaojie and Ling, Haibin},
  journal={TPAMI},
  year={2020}
}

@inproceedings{liang2022fusion,
  title={Fusion from decomposition: A self-supervised decomposition approach for image fusion},
  author={Liang, Pengwei and Jiang, Junjun and Liu, Xianming and Ma, Jiayi},
  booktitle={ECCV},
  year={2022}
}

@article{liu2022attention,
  title={Attention-guided global-local adversarial learning for detail-preserving multi-exposure image fusion},
  author={Liu, Jinyuan and Shang, Jingjie and Liu, Risheng and Fan, Xin},
  journal={TCSVT},
  year={2022}
}

@article{ma2022end,
  title={End-to-end learning for simultaneously generating decision map and multi-focus image fusion result},
  author={Ma, Boyuan and Yin, Xiang and Wu, Di and Shen, Haokai and Ban, Xiaojuan and Wang, Yu},
  journal={Neurocomputing},
  year={2022}
}

@inproceedings{guan2023mutual,
  title={Mutual-guided dynamic network for image fusion},
  author={Guan, Yuanshen and Xu, Ruikang and Yao, Mingde and Wang, Lizhi and Xiong, Zhiwei},
  booktitle={ACMMM},
  year={2023}
}

@article{zhang2024exploit,
title={Exploit the best of both end-to-end and map-based methods for multi-focus image fusion},
author={Zhang, Juncheng and Liao, Qingmin and Ma, Haoyu and Xue, Jing-Hao and Yang, Wenming and Liu, Shaojun},
journal={TMM},
year={2024},
}

@inproceedings{krizhevsky2012imagenet,
  title={Imagenet classification with deep convolutional neural networks},
  author={Krizhevsky, Alex and Sutskever, Ilya and Hinton, Geoffrey E},
  booktitle={NeurIPS},
  year={2012}
}

@inproceedings{xie2022pyramid,
  title={Pyramid grafting network for one-stage high resolution saliency detection},
  author={Xie, Chenxi and Xia, Changqun and Ma, Mingcan and Zhao, Zhirui and Chen, Xiaowu and Li, Jia},
  booktitle={CVPR},
  year={2022}
}

@article{tang2022piafusion,
  title={Piafusion: A progressive infrared and visible image fusion network based on illumination aware},
  author={Tang, Linfeng and Yuan, Jiteng and Zhang, Hao and Jiang, Xingyu and Ma, Jiayi},
  journal={Information Fusion},
  year={2022}
}

@inproceedings{liu2022target,
  title={Target-aware dual adversarial learning and a multi-scenario multi-modality benchmark to fuse infrared and visible for object detection},
  author={Liu, Jinyuan and Fan, Xin and Huang, Zhanbo and Wu, Guanyao and Liu, Risheng and Zhong, Wei and Luo, Zhongxuan},
  booktitle={CVPR},
  year={2022}
}

@inproceedings{xu2020fusiondn,
  title={Fusiondn: A unified densely connected network for image fusion},
  author={Xu, Han and Ma, Jiayi and Le, Zhuliang and Jiang, Junjun and Guo, Xiaojie},
  booktitle={AAAI},
  year={2020}
}

@article{zhang2023visible,
  title={Visible and infrared image fusion using deep learning},
  author={Zhang, Xingchen and Demiris, Yiannis},
  journal={TPAMI},
  year={2023}
}

@inproceedings{zhao2023metafusion,
  title={Metafusion: Infrared and visible image fusion via meta-feature embedding from object detection},
  author={Zhao, Wenda and Xie, Shigeng and Zhao, Fan and He, You and Lu, Huchuan},
  booktitle={CVPR},
  year={2023}
}

@inproceedings{zhao2023cddfuse,
  title={Cddfuse: Correlation-driven dual-branch feature decomposition for multi-modality image fusion},
  author={Zhao, Zixiang and Bai, Haowen and Zhang, Jiangshe and Zhang, Yulun and Xu, Shuang and Lin, Zudi and Timofte, Radu and Van Gool, Luc},
  booktitle={CVPR},
  year={2023}
}

@article{li2023lrrnet,
  title={Lrrnet: A novel representation learning guided fusion network for infrared and visible images},
  author={Li, Hui and Xu, Tianyang and Wu, Xiao-Jun and Lu, Jiwen and Kittler, Josef},
  journal={TPAMI},
  year={2023}
}

@article{xu2023murf,
  title={Murf: Mutually reinforcing multi-modal image registration and fusion},
  author={Xu, Han and Yuan, Jiteng and Ma, Jiayi},
  journal={TPAMI},
  year={2023}
}

@inproceedings{zhao2023ddfm,
  title={Ddfm: Denoising diffusion model for multi-modality image fusion},
  author={Zhao, Zixiang and Bai, Haowen and Zhu, Yuanzhi and Zhang, Jiangshe and Xu, Shuang and Zhang, Yulun and Zhang, Kai and Meng, Deyu and Timofte, Radu and Van Gool, Luc},
  booktitle={ICCV},
  year={2023}
}

@inproceedings{liu2023multi,
  title={Multi-interactive feature learning and a full-time multi-modality benchmark for image fusion and segmentation},
  author={Liu, Jinyuan and Liu, Zhu and Wu, Guanyao and Ma, Long and Liu, Risheng and Zhong, Wei and Luo, Zhongxuan and Fan, Xin},
  booktitle={ICCV},
  year={2023}
}

@article{tang2023divfusion,
  title={Divfusion: Darkness-free infrared and visible image fusion},
  author={Tang, Linfeng and Xiang, Xinyu and Zhang, Hao and Gong, Meiqi and Ma, Jiayi},
  journal={Information Fusion},
  year={2023}
}

@inproceedings{he2016deep,
  title={Deep residual learning for image recognition},
  author={He, Kaiming and Zhang, Xiangyu and Ren, Shaoqing and Sun, Jian},
  booktitle={CVPR},
  year={2016}
}

@article{zhang2021image,
  title={Image fusion meets deep learning: A survey and perspective},
  author={Zhang, Hao and Xu, Han and Tian, Xin and Jiang, Junjun and Ma, Jiayi},
  journal={Information Fusion},
  year={2021}
}

@inproceedings{zhang2024mrfs,
  title={Mrfs: Mutually reinforcing image fusion and segmentation},
  author={Zhang, Hao and Zuo, Xuhui and Jiang, Jie and Guo, Chunchao and Ma, Jiayi},
  booktitle={CVPR},
  year={2024}
}

@article{li2024crossfuse,
  title={Crossfuse: A novel cross attention mechanism based infrared and visible image fusion approach},
  author={Li, Hui and Wu, Xiao-Jun},
  journal={Information Fusion},
  year={2024}
}

@article{zhang2024text,
  title={Text-difuse: An interactive multi-modal image fusion framework based on text-modulated diffusion model},
  author={Zhang, Hao and Cao, Lei and Ma, Jiayi},
  booktitle={NeurIPS},
  year={2024}
}
}
\clearpage
\setcounter{page}{1}
\maketitlesupplementary

\setcounter{table}{0}
\setcounter{figure}{0}
\setcounter{section}{0}
\setcounter{equation}{0}
\renewcommand{\thetable}{A\arabic{table}}
\renewcommand{\thefigure}{A\arabic{figure}}
\renewcommand{\thesection}{A\arabic{section}}
\renewcommand{\theequation}{A\arabic{equation}}

\section{More Qualitative Comparison for Multi-Exposure Image Fusion}\label{sec:suppl_mef}
\begin{figure*}[t]
    \centering
    \includegraphics[width=\linewidth]{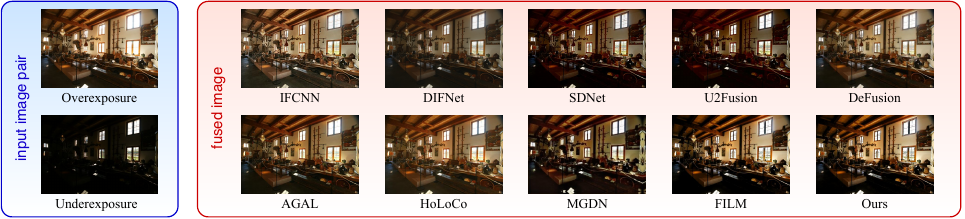}
    \caption{Visualization comparison of fusion results on the MEFB~\cite{zhang2021benchmarking} dataset for the multi-exposure image fusion task.}
    \label{fig:Qualitative Comparison on MEFB}
\end{figure*}

We provide additional qualitative results for multi-exposure image fusion (MEF). As shown in Fig.~\ref{fig:Qualitative Comparison on MEFB}, most prior methods fail to maintain appropriate luminance. AGAL~\cite{liu2022attention}, HoLoCo~\cite{liu2023holoco}, and FILM~\cite{Zhao_2024_ICML} produce results suffering from considerable detail loss in high-intensity regions. In contrast, our method consistently preserves optimal brightness and contrast, while effectively retaining structural and textural information, such as the details of grass and trees outside the window, thereby delivering perceptually superior fusion outputs. This demonstrates the effectiveness and robustness of our method, where the multi-grained textual guidance modulates visual features at corresponding levels, and the hierarchical supervision together with the saliency-driven visual enrichment module, further improve cross-modal feature alignment and auxiliary text utilization.

\section{More Qualitative Comparison for Multi-Focus Image Fusion}\label{sec:suppl_mff}
\begin{figure*}[t]
    \centering
    \includegraphics[width=\linewidth]{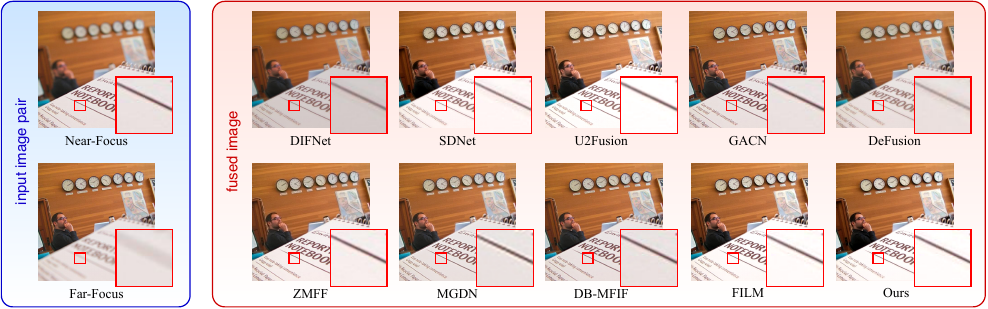}
    \caption{Visualization comparison of fusion outputs on the Lytro~\cite{nejati2015multi} dataset for the multi-focus image fusion task.}
    \label{fig:Qualitative Comparison on Lytro}
\end{figure*}

We present additional qualitative results for multi-focus image fusion (MFF). As shown in Fig.~\ref{fig:Qualitative Comparison on Lytro}, most compared methods suffer from insufficient detail preservation and noticeable artifacts. U2Fusion~\cite{xu2020u2fusion} and FILM~\cite{Zhao_2024_ICML} suffer from overexposure, leading to unnatural appearances in facial regions. In contrast, our method consistently generates visually balanced outputs with enhanced detail, natural luminance, and artifact-free quality, as demonstrated by the clear text and distinct lines on the book page. These results demonstrate the effectiveness and robustness of our method in preserving focused details and avoiding artifacts in MFF.

\section{Infrared and Visible Image Fusion}\label{sec:suppl_ivf}
To further validate the generalizability and robustness of our proposed \textbf{M}ulti-grained \textbf{T}ext-guided \textbf{I}mage \textbf{F}usion (MTIF) framework, we extend its application from multi-exposure image fusion (MEF) and multi-focus image fusion (MFF) tasks to the infrared and visible image fusion (IVF). IVF aims to synthesize a single image that integrates thermal radiation cues from infrared sensors and fine-grained texture and color details from RGB cameras, enabling more comprehensive scene perception under adverse conditions such as low illumination, smoke, or haze~\cite{tang2023divfusion}. Despite its benefits, IVF remains a challenging task due to the significant modality gap between infrared and visible images, including disparities in spectral characteristics, spatial structures, and semantic content~\cite{zhang2023visible}.

\noindent\textbf{Dataset and Setup.}
Following Zhao et al.\cite{Zhao_2024_ICML}, we use a subset of the MSRS\cite{tang2022piafusion} dataset for training, with the remainder reserved for testing. This dataset includes a wide variety of image pairs from different scenes. To further assess the generalizability of MTIF, we conduct experiments on additional datasets, including M3FD~\cite{liu2022target}, which covers four major scenarios with various environments, illumination,
seasons, and weather, and RoadScene~\cite{xu2020fusiondn}, which contains road scene images captured in various outdoor settings. All training settings, including the number of epochs, optimization strategy, and model configurations, are consistent with those used in the MEF experiments.
For the IVF task, we compare our method with nine state-of-the-art deep learning methods: TarDAL~\cite{liu2022target}, DeFusion~\cite{liang2022fusion}, MetaFusion~\cite{zhao2023metafusion}, CDDFuse~\cite{zhao2023cddfuse}, LRRNet~\cite{li2023lrrnet}, MURF~\cite{xu2023murf}, DDFM~\cite{zhao2023ddfm}, SegMIF~\cite{liu2023multi}, and FILM~\cite{Zhao_2024_ICML}.

\begin{figure}[!ht]
    \centering
    \includegraphics[width=\linewidth]{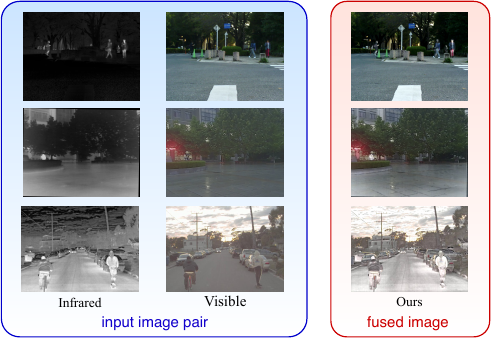}
    \caption{Visualization comparison of fusion results for the infrared-visible image fusion task.}
    \label{fig:Qualitative Comparison for IVF}
\end{figure}

\noindent\textbf{Results and Analysis.}
We present both qualitative and quantitative evaluations to assess the performance of our method. For the qualitative evaluation, Fig.~\ref{fig:Qualitative Comparison for IVF} compares the input images with the fusion outputs generated by our method. The results clearly demonstrate that our method effectively preserves thermal radiation and fine details while minimizing artifacts through multi-grained textual guidance. For the quantitative results in Tab.~\ref{tab:IVF}, our method achieves superior performance in most metrics, demonstrating its effectiveness in preserving thermal and fine details and minimizing artifacts in diverse IVF tasks.

    \begin{table*}[!htb]
        \centering
        \caption{Quantitative results of infrared and visible image fusion. The \textbf{bold} markers represent the best values.}
        \label{tab:IVF}%
        \resizebox{\linewidth}{!}{
            \begin{tabular}{crrrrrrcrrrrrrcrrrrrr}
                \toprule
                \multicolumn{7}{c}{\textbf{MSRS Infrared-Visible Fusion Dataset}} & \multicolumn{7}{c}{\textbf{M$^3$FD Infrared-Visible Fusion Dataset}} & \multicolumn{7}{c}{\textbf{RoadScene Infrared-Visible Fusion Dataset}} \\
                & \multicolumn{1}{c}{EN} & \multicolumn{1}{c}{SD} & \multicolumn{1}{c}{SF} & \multicolumn{1}{c}{AG} & \multicolumn{1}{c}{VIF} & \multicolumn{1}{c}{Qabf} &       & \multicolumn{1}{c}{EN} & \multicolumn{1}{c}{SD} & \multicolumn{1}{c}{SF} & \multicolumn{1}{c}{AG} & \multicolumn{1}{c}{VIF} & \multicolumn{1}{c}{Qabf} &       & \multicolumn{1}{c}{EN} & \multicolumn{1}{c}{SD} & \multicolumn{1}{c}{SF} & \multicolumn{1}{c}{AG} & \multicolumn{1}{c}{VIF} & \multicolumn{1}{c}{Qabf} \\
                \midrule
                TarD~\cite{liu2022target}  & 5.28  & 25.22 & 5.98  & 1.83  & 0.42  & 0.18  & TarD~\cite{liu2022target}  & 6.79  & 40.75 & 8.18  & 2.92  & 0.53  & 0.30  & TarD~\cite{liu2022target}  & 7.25  & 47.57 & 11.46 & 4.23  & 0.56  & 0.43 \\
                DeF~\cite{liang2022fusion}   & 6.46  & 37.63 & 8.60  & 2.80  & 0.77  & 0.54  & DeF~\cite{liang2022fusion}   & 6.84  & 35.09 & 9.65  & 3.37  & 0.59  & 0.42  & DeF~\cite{liang2022fusion}   & 7.39  & 47.60 & 11.26 & 4.47  & 0.63  & 0.50 \\
                Meta~\cite{zhao2023metafusion}  & 5.65  & 24.97 & 9.99  & 3.40  & 0.63  & 0.48  & Meta~\cite{zhao2023metafusion}  & 6.68  & 29.62 & 16.22 & 5.68 & 0.68  & 0.57  & Meta~\cite{zhao2023metafusion}  & 6.87  & 31.95 & 14.40 & 5.55  & 0.55  & 0.46 \\
                CDDF~\cite{zhao2023cddfuse}  & 6.70 & \textbf{43.39} & 11.56 & 3.74 & 1.05 & 0.69 & CDDF~\cite{zhao2023cddfuse}  & 7.08 & 41.29 & 16.49 & 5.42  & 0.78 & 0.63  & CDDF~\cite{zhao2023cddfuse}  & 7.41 & \textbf{54.59} & 17.04 & 6.07 & 0.63  & 0.51 \\
                LRR~\cite{li2023lrrnet}   & 6.19  & 31.78 & 8.46  & 2.63  & 0.54  & 0.46  & LRR~\cite{li2023lrrnet}   & 6.60  & 30.19 & 11.69 & 3.95  & 0.57  & 0.51  & LRR~\cite{li2023lrrnet}   & 7.09  & 38.77 & 11.50 & 4.36  & 0.43  & 0.33 \\
                MURF~\cite{xu2023murf}  & 5.04  & 20.63 & 10.49 & 3.38  & 0.44  & 0.36  & MURF~\cite{xu2023murf}  & 6.52  & 27.90 & 11.43 & 4.51  & 0.39  & 0.30  & MURF~\cite{xu2023murf}  & 6.91  & 33.46 & 13.74 & 5.31  & 0.53  & 0.47 \\
                DDFM~\cite{zhao2023ddfm}  & 6.19  & 29.26 & 7.44  & 2.51  & 0.73  & 0.48  & DDFM~\cite{zhao2023ddfm}  & 6.72  & 31.15 & 9.84  & 3.42  & 0.63  & 0.47  & DDFM~\cite{zhao2023ddfm}  & 7.27  & 42.94 & 10.89 & 4.20  & 0.63  & 0.50 \\
                SegM~\cite{liu2023multi}  & 5.95  & 37.28 & 11.10 & 3.47  & 0.88  & 0.63  & SegM~\cite{liu2023multi}  & 6.89  & 35.64 & 16.11 & 5.52  & 0.78  & 0.65 & SegM~\cite{liu2023multi}  & 7.29  & 47.10 & 15.07 & 5.78  & 0.65 & 0.56 \\
                FILM~\cite{Zhao_2024_ICML}  & 6.72 & 43.17 & 11.70 & 3.84 & 1.06 & 0.73 & FILM~\cite{Zhao_2024_ICML}  & 7.09 & 41.53 & 16.77 & 5.55 & 0.83 & 0.67 & FILM~\cite{Zhao_2024_ICML}  & 7.43 & 49.25 & 17.34 & 6.60 & \textbf{0.69} & 0.62 \\
                Ours  & \textbf{6.73} & 43.31 & \textbf{11.72} & \textbf{3.84} & \textbf{1.06} & \textbf{0.73} & Ours  & \textbf{7.15} & \textbf{42.88} & \textbf{17.08} & \textbf{5.77} & \textbf{0.86} & \textbf{0.69} & Ours  & \textbf{7.53} & 54.13 & \textbf{18.14} & \textbf{6.92} & 0.68 & \textbf{0.62} \\
                \bottomrule
        \end{tabular}}
    \end{table*}%
\begin{table*}[!ht]
    \centering
    \renewcommand\arraystretch{1.0}
    \caption{Ablation of hyper-parameters on multi-exposure and multi-focus image fusion tasks. The \textbf{bold} markers represent the best values.}
    \label{tab:ablation_hyper-parameters}
    \setlength{\tabcolsep}{10pt}
    \resizebox{0.9\linewidth}{!}{
    \begin{tabular}{l|cccccc|cccccc}
        \toprule
        \multirow{2}{*}{Configuration} 
        & \multicolumn{6}{c|}{\textbf{SICE Multi-exposure Image Fusion Dataset}} 
        & \multicolumn{6}{c}{\textbf{RealMFF Multi-focus Image Fusion Dataset}} \\
        \cmidrule(lr){2-7} \cmidrule(lr){8-13}
        & EN & SD & SF & AG & VIF & Qabf 
        & EN & SD & SF & AG & VIF & Qabf \\
        \midrule
        $\alpha_{1}$ = 20
        & 7.28  & 56.38 & 18.88 & \textbf{5.52}  & 1.20  & 0.77 
        & 7.13  & 56.04 & 15.72 & 5.40  & \textbf{1.59}  & 0.76 \\
        
        $\alpha_{1}$ = 5 
        & 7.26  & 59.52 & 18.81 & 5.43  & 1.43  & 0.76 
        & 7.13  & 55.30 & 15.45 & 5.28  & 1.58  & 0.76 \\
        
        $\alpha_{2}$ = 5 
        & 7.25  & 59.54 & 18.64 & 5.37  & 1.44  & 0.77 
        & 7.11  & 55.87 & 15.55 & 5.31  & 1.58  & 0.76 \\

        $\alpha_{2}$ = 0.2 
        & 7.28  & 60.06 & 18.72 & 5.43  & 1.46  & 0.77 
        & 7.13  & \textbf{57.39} & 15.59 & 5.39  & 1.59  & 0.75 \\

        $\beta_{1}$ = 2 
        & 7.13  & 58.30 & \textbf{19.24} & 5.50  & 1.42  & 0.75 
        & 7.13  & 56.01 & 15.34 & 5.26  & 1.58  & 0.76 \\

        $\beta_{1}$ = 0.5 
        & 7.26  & 59.69 & 18.81 & 5.41  & 1.46  & 0.77 
        & 7.13  & 56.73 & 15.69 & 5.36  & 1.58  & 0.76 \\

        $\beta_{2}$ = 400 
        & 7.27  & 59.11 & 18.82 & 5.43  & 1.44  & 0.76 
        & 7.12  & 55.31 & 15.49 & 5.29  & 1.58  & 0.76 \\

        $\beta_{2}$ = 25 
        & 7.25  & 59.49 & 18.84 & 5.47  & 1.45  & 0.76 
        & 7.11  & 54.83 & 15.56 & 5.31  & 1.57  & 0.76 \\
      
        \midrule
        \textbf{Ours}
        & \textbf{7.28} & \textbf{60.41} & 18.78 & 5.40 & \textbf{1.46} & \textbf{0.79}
        & \textbf{7.14} & 56.03 & \textbf{15.78} & \textbf{5.43} & 1.56 & \textbf{0.77} \\
        \bottomrule
    \end{tabular}}
\end{table*}

\section{Evaluation of Hyper-Parameters}
We perform a hyper-parameter analysis of the loss function to assess the effects of different weighting strategies. As shown in Tab.~\ref{tab:ablation_hyper-parameters}, excessive weighting of $\alpha_{1}$ or $\alpha_{2}$ degrades fine detail preservation. Increasing $\beta_{1}$ slightly decreases contrast, while decreasing it reduces global information preservation. For $\beta_{2}$, overweighting compromises structural perception, whereas insufficient weighting underutilizes saliency, impairing overall visual quality. The proposed default configuration achieves an optimal balance and consistently achieves optimal performance on both multi-exposure image fusion and multi-focus image fusion tasks.

\end{document}